\documentclass[pmlr]{jmlr}

\usepackage{longtable}

\usepackage{booktabs}
\usepackage[load-configurations=version-1]{siunitx} 

\theorembodyfont{\upshape}
\theoremheaderfont{\scshape}
\theorempostheader{:}
\theoremsep{\newline}

\jmlrvolume{133}
\jmlryear{2021}
\jmlrworkshop{
NeurIPS 2020 Competition and Demonstration Track
}

\def\firstpageno#1{\setcounter{page}{#1}}
\firstpageno{86}
 
\newif\ifsqueeze
\squeezefalse 

\usepackage{multirow}

\newcommand{\low}{500MiB}
\newcommand{\medium}{6GiB}
\newcommand{\smallacc}{25\%}

\newcommand{\msunrestricted}{MS UnitedQA}
\newcommand{\fbunrestricted}{FB Hybrid}
\newcommand{\fbmedium}{FB system}
\newcommand{\soseki}{Ousia-Tohoku Soseki}
\newcommand{\brno}{BUT R2-D2}
\newcommand{\ucl}{UCLNLP-FB system}
\newcommand{\naver}{NAVER RDR}
\newcommand{\uclsmall}{UCLNLP-FB system (29M)}

\usepackage[disable]{todonotes}
\newcommand{\tk}[1]{\todo[color=magenta]{TK: #1}} 
\newcommand{\jbg}[1]{\todo[color=cyan]{JBG: #1}}  
\newcommand{\danqi}[1]{\todo[color=teal!20]{Danqi: #1}}  
\newcommand{\ec}[1]{\todo[color=magenta]{EC: #1}} 
\newcommand{\sm}[1]{\todo[color=BurntOrange]{Sewon: #1}} 

\newcommand{\red}[1]{{\color{red} #1}}
\newcommand{\blue}[1]{{\color{blue} #1}}
\newcommand{\white}[1]{{\color{white} #1}}
\newcommand{\stat}[2]{(\red{#1\%}, \blue{#2\%})}

\newcommand{\affilsup}[1]{\rlap{\textsuperscript{\normalfont#1}}}

\newcommand{\taffilsup}[1]{\affilsup{T#1}\hspace{.2cm}}
\newcommand{\taffilsupwide}[1]{\affilsup{T#1}\hspace{.45cm}}

\newcommand{\participants}{Patrick Lewis\taffilsup{1},
    Yuxiang Wu\taffilsup{1},
    Heinrich Küttler\taffilsup{1},
    Linqing Liu\taffilsup{1},   \\
    Pasquale Minervini\taffilsup{1},
    Pontus Stenetorp\taffilsup{1},
    Sebastian Riedel\taffilsupwide{1,3},
    Sohee Yang\taffilsup{2}, \\
    Minjoon Seo\taffilsup{2},
    Gautier Izacard\taffilsupwide{3,7},
    Fabio Petroni\taffilsupwide{3,7},
    Lucas Hosseini\taffilsupwide{3,7}, \\
    Nicola De Cao\taffilsup{3},
    Edouard Grave\taffilsupwide{3,7},
    Ikuya Yamada\taffilsup{4},
    Sonse Shimaoka\taffilsup{4}, \\
    Masatoshi Suzuki\taffilsup{4},
    Shumpei Miyawaki\taffilsup{4},
    Shun Sato\taffilsup{4},
    Ryo Takahashi\taffilsup{4}, \\
    Jun Suzuki\taffilsup{4},
    Martin Fajcik\taffilsup{5},
    Martin Docekal\taffilsup{5},
    Karel Ondrej\taffilsup{5},
    Pavel Smrz\taffilsup{5}, \\
    Hao Cheng\taffilsup{6},
    Yelong Shen\taffilsup{6},
    Xiaodong Liu\taffilsup{6},
    Pengcheng He\taffilsup{6}, 
    Weizhu Chen\taffilsup{6}, \\
    Jianfeng Gao\taffilsup{6},
    Barlas Oguz\taffilsup{7},
    Xilun Chen\taffilsup{7},
    Vladimir Karpukhin\taffilsup{7}, \\
    Stan Peshterliev\taffilsup{7},
    Dmytro Okhonko\taffilsup{7},
    Michael Schlichtkrull\taffilsup{7},
    Sonal Gupta\taffilsup{7}, \\
    Yashar Mehdad\taffilsup{7},
    Wen-tau Yih\taffilsup{7}
}

\title[NeurIPS 2020 EfficientQA competition]{NeurIPS 2020 EfficientQA Competition: \titlebreak Systems, Analyses and Lessons Learned}

 \author{
 \Name{Sewon Min\nametag{$^2$}} \Email{sewon@cs.washington.edu} \AND
 \Name{Jordan Boyd-Graber\nametag{$^3$}} \Email{jbg@umiacs.umd.edu} \AND
 \Name{Chris Alberti\nametag{$^1$}} \Email{chrisalberti@google.com} \AND
 \Name{Danqi Chen\nametag{$^4$}} \Email{danqic@cs.princeton.edu} \AND
 \Name{Eunsol Choi\nametag{$^5$}} \Email{eunsol@cs.utexas.edu} \AND
 \Name{Michael Collins\nametag{$^1$}} \Email{mjcollins@google.com} \AND
 \Name{Kelvin Guu\nametag{$^1$}} \Email{kguu@google.com} \AND
 \Name{Hannaneh Hajishirzi\nametag{$^1$}} \Email{hannaneh@cs.washington.edu} \AND
 \Name{Kenton Lee\nametag{$^1$}} \Email{kentonl@google.com} \AND
 \Name{Jennimaria Palomaki\nametag{$^1$}} \Email{jpalomaki@google.com} \AND
 \Name{Colin Raffel\nametag{$^1$}} \Email{craffel@google.com} \AND
 \Name{Adam Roberts\nametag{$^1$}} \Email{adarob@google.com} \AND
 \Name{Tom Kwiatkowski\nametag{$^1$}} \Email{tomkwiat@google.com} \vspace{.5em} \\
 \addr {$^1$Google Research, New York City, NY, USA; Mountain View, CA, USA; Seattle, WA, USA \\
    $^2$University of Washington, Seattle, WA, USA \\
    $^3$University of Maryland, College Park, MD, USA \\
    $^4$Princeton University, Princeton, NJ, USA \\
    $^5$University of Texas at Austin, Austin, TX, USA
  }
  \vspace{1.5em} \\ 
  \Name{\participants} \vspace{.5em} \\
  \addr{$^{\mathrm{T}1}$UCLNLP \& Facebook AI;
        $^{\mathrm{T}2}$NAVER Clova;
        $^{\mathrm{T}3}$Facebook AI Paris \& London; \\
        $^{\mathrm{T}4}$Studio Ousia, Tohoku University \& RIKEN;
        $^{\mathrm{T}5}$Brno University of Technology; \\
        $^{\mathrm{T}6}$Microsoft Research \& Dynamic 365 AI;
        $^{\mathrm{T}7}$Facebook AI}
 }

\editors{Hugo Jair Escalante and Katja Hofmann}

\begin{document}

\maketitle

\begin{abstract}
We review the EfficientQA competition\footnote{http://efficientqa.github.io/} from NeurIPS 2020\footnote{https://neurips.cc/Conferences/2020/CompetitionTrack}.
The competition focused on open-domain question answering (QA), where systems take natural language questions as input and return natural language answers.
\danqi{Two ``natural language'' --- feels duplicated}
\tk{I prefer to be explicit to set this apart from e.g. semantic parsing tasks where the input is natural language, but the output is constrained by the contents of a database.}
The aim of the competition was to build systems that can predict correct answers while also satisfying strict on-disk memory budgets.
These memory budgets were designed to encourage contestants to explore the trade-off between storing retrieval corpora or the parameters of learned models.
In this report, we describe the motivation and organization of the competition, review the best submissions, and analyze system predictions to inform a discussion of evaluation  for open-domain QA.
\end{abstract}

\section{Introduction}
Open-domain question answering (QA) is emerging as a benchmark method of measuring computational systems' abilities to retrieve, represent, and read knowledge~\citep{voorhees-00, chen2017reading, seo2019real, lee2019latent}.
Recently, this task has been addressed by a diverse set of approaches that navigate multiple documents~\citep{min2019knowledge, asai2019learning}, index large corpora of text~\citep{lee2019latent, guu2020realm, karpukhin2020dense}, represent world knowledge in the parameters of a neural network~\citep{roberts2020knowledge}, or consolidate knowledge from multiple passages~\citep{izacard2020leveraging}.
More comprehensive background is provided in \citet{chen2020open}.

The EfficientQA competition, held at NeurIPS 2020, required contestants to build self-contained systems that contain all of the knowledge required to answer open-domain questions.
There were no constraints on how the knowledge is stored---it could be in documents, databases, the parameters of a neural network, or any other form.
However, the competition encouraged systems that store and access this knowledge using the smallest number of bytes, including code, corpora, and model parameters.
Specifically, EfficientQA had four tracks: 1) best accuracy overall (unconstrained); 2) best accuracy, system size under \medium{}; 3) best accuracy, system size under \low{}; 4) smallest system to get \smallacc{} accuracy.
These memory budgets were designed to encourage contestants to explore the trade-off between storing and accessing large, redundant, retrieval corpora, structured data stores, or the parameters of large learned models.

This paper summarizes the findings from the competition.
Section~\ref{sec:details} describes the competition in detail; Section~\ref{sec:systems} presents a description of the best performing submissions in each track; Section~\ref{sec:annotation} introduces a new human evaluation of system accuracy, and Section~\ref{sec:results-and-analysis} provides results and an analysis based on automatic and human evaluation.
Finally, Section~\ref{sec:trivia} pits the top systems against human trivia experts.

\subsection{Key takeaways}
The top submissions in each of EfficientQA's four tracks significantly outperformed the provided baselines. 
All top submissions use a retrieval corpus and a neural network answering module.
However, the nature of the retrieval corpus and answering module differs drastically across the tracks (Table~\ref{tab:system-summary}).

\begin{table}[t]
\begin{center} \setlength\tabcolsep{3.5pt} \scriptsize 
\begin{tabular}{l|l|l|c|c|c}
\toprule
    Track & Model & Affiliation & retr & answer & others \\
\midrule
    & REALM & Organizers      & $p$ & ext &  \\
    Unrest- & DPR & Organizers        & $p$ & ext & \\
    ricted & \msunrestricted & Microsoft \& Dynamics 365 & $p$ & ext+gen & \\
    & \fbunrestricted & Facebook AI & $p$ & gen & lists/tables \\
\midrule
    & DPR-subset & Organizers & $p$ (pruned) & ext & \\
    & T5-XL+SSM & Organizers & X & gen \\
    \medium & \fbmedium & FAIR-Paris\&London & $p$ (pruned) & gen & lists, lrzip compression \\
    & \soseki & Studio Ousia, Tohoku U \& RIKEN & $p$ (pruned) & ext & ZPAQ compression \\
    & \brno & Brno U of Technology & $p$ (pruned) & ext+gen \\
\midrule
    & T5-Small+SSM & Organizers & X & gen \\
    \low & \ucl & UCL \& FAIR & ($q$,$a$) & - & data augmentation \\
    & \naver & NAVER Clova & $p$ (pruned) & ext & single Transformer \\
\midrule
    \smallacc\ & T5-XL+SSM & Organizers & X & gen \\
    smallest & \uclsmall & UCL \& FAIR & ($q$,$a$) & - & data augmentation \\
\bottomrule
\end{tabular}
\caption{A list of the baselines and systems from participants, along with the team affiliation and key distinction between systems.
{\em retr} means what the systems retrieves, e.g., passages ($p$), question-answer pairs (($q$, $a$)), or none (X).
{\em pruned} indicates the Wikipedia corpus is pruned.
{\em answer} indicates whether the answer is extracted ({\em ext}) or generated ({\em gen}).
Other key distinctions are shown in the last column.
}
\label{tab:system-summary}
\end{center}
\end{table}

\paragraph{Unrestricted and 6GiB tracks} The top submissions to the unrestricted track and the \medium\ track (Section~\ref{subsec:unconstrained-track}; \ref{subsec:medium-track}) outperformed the state-of-the-art baselines from April 2020 by nearly 20\%.
They achieved this improvement by combining the state-of-the-art retrieval systems~\citep{karpukhin2020dense,GAR} with answer generation \citep{izacard2020leveraging}; leveraging the state-of-the-art in text generation \citep{t5} and text encoding \citep{clark2020electra}; modeling not only text but also tables and lists from Wikipedia; and combining the extractive and generative answer prediction.
The top submissions to the \medium\ track additionally massively reduced the size of their indexed corpus and made use of the state-of-the-art in compression, with minimal impact on 
accuracy.

\paragraph{500MiB and smallest tracks} To get under 500MB (Section~\ref{subsec:low-track}), the systems made more drastic changes.
The submission from NAVER Clova drastically reduced the size of their indexed corpus and reused a single Transformer model for the retriever and the reader, winning the \low\ track according to the human evaluation.
The even smaller \ucl\ took a novel approach in generating a large corpus of question-answer pairs, indexing it, and retrieving the most similar question to the input question.
This approach, with two systems with different sizes in question-answer corpus, won both the \low\ track and the smallest \smallacc\ track according to the automatic evaluation.

\paragraph{Automatic vs. human evaluation}
The human evaluation supports the observation that automatic metrics often incorrectly penalize correct predictions in the open-domain setting \citep{voorhees-00, roberts2020knowledge}.
We also investigate the effect of question ambiguity on evaluation---the questions from NQ are often ambiguous without the associated evidence document \citep{min2020ambigqa}.
In Section~\ref{sec:annotation} we define an annotation scheme that supports multiple estimations of accuracy, corresponding to different definitions of correctness for answers to ambiguous questions.
Almost all systems' accuracy increased by 20\%-25\% under the strictest definition of correctness.
The increase doubled when we relaxed the definition of correctness to permit any semantically valid interpretation of the question.
We present a discussion 
as well as suggestions for future evaluation (Section~\ref{sec:analysis}).


\section{Competition Overview}\label{sec:details}

\paragraph{Data}
The competition uses English questions and answers from the Natural Questions dataset~\cite[NQ]{kwiatkowski2019natural}: real user questions issued to the Google search engine, along with reference answers from Wikipedia\footnote{NQ was collected over the course of 2017 and 2018. Most solutions to the open-domain variant use a Wikipedia dump from December 2018. See \cite{lee2019latent} for details.}.
Real user questions in NQ are interpretable outside of a document context (unlike SQuAD~\citep{rajpurkar-16}), and less amenable to traditional IR approaches than over-complete trivia questions~\citep{joshi-17}, as discussed in \citet{lee2019latent}.
While the original NQ task was posed as a reading comprehension task, in which evidence documents are provided, recent work has adapted the NQ data to the open-domain setting \citep{lee2019latent, min2019discrete, min2019knowledge, asai2019learning, guu2020realm, roberts2020knowledge, karpukhin2020dense} by taking examples with short answers (up to five tokens) and discarding evidence documents.
In the open-domain setting, NQ contains 88k one-way annotated training examples and 4k five-way annotated development examples.
For EfficientQA, we introduce a new test and development set constructed in the same way as the original NQ, but labeled slightly after (early 2019 rather than through 2018).\footnote{This temporal change caused slight degradation in performance for baseline systems.}
Our test set was kept hidden from contestants, and submissions were made by uploading solutions to an automatic leaderboard. 

\paragraph{System size}
All submissions to the three restricted tracks were submitted as self-contained Docker\footnote{https://www.docker.com} images.
We defined system size to be the on-disk, at-rest size of this image.
We chose this approach to avoid confusion about distinctions between data, model parameters, and code.
\ifsqueeze
\else
    However, this choice did lead to a significant engineering efforts for the very smallest systems, which were not able to build on the standard Docker templates for the predominant deep-learning libraries.
    While the organizers did provide a small reference system based on T5 \citep{roberts2020knowledge} that used TensorFlow serving\footnote{https://www.tensorflow.org/tfx/serving/docker}, this did not support most submissions, and the very smallest systems required clever compilation strategies on top of their modeling enhancements.
\fi

\paragraph{Evaluation metrics}
We measured the performance of different systems through automatic and human evaluation.
For automatic evaluation, the accuracy of each systems' predicted answers is judged against reference annotations, annotated by five human workers.
We follow the literature in using exact match between predicted and reference answers after minor normalization \citep{lee2019latent}.
Due to the ambiguities inherent in language and question-answering in general, five reference answers are often not exhaustive, and systems predict correct answers that are judged incorrect according to the automatic metrics.
To rectify this, we sent predictions from each of the systems for further human rating by three
raters, to get a better estimation of accuracy, as detailed in Section~\ref{sec:annotation}. 

\paragraph{Competition schedule}
The competition was announced in June 2020, along with baselines and tutorials.
The official leaderboard for restricted settings was launched on September 14th, 2020, and participants had two months to submit their systems (November 14th, 2020).
Finally, predictions from the top systems from each leaderboard were sent for human evaluation, which was completed in the end of November.
The same set of top systems was invited to present their systems at the NeurIPS event on December 12th and in this paper.
The human vs.\ computer competition was held on December 6th (details in Section~\ref{sec:trivia}).

In total, we had 39 submissions from 18 unique teams, seven of which were affiliated or co-affiliated with universities.


\section{Systems}\label{sec:systems}
\begin{figure} 
  \vspace{-1.5em}
  \begin{center}
    \includegraphics[width=0.85\textwidth]{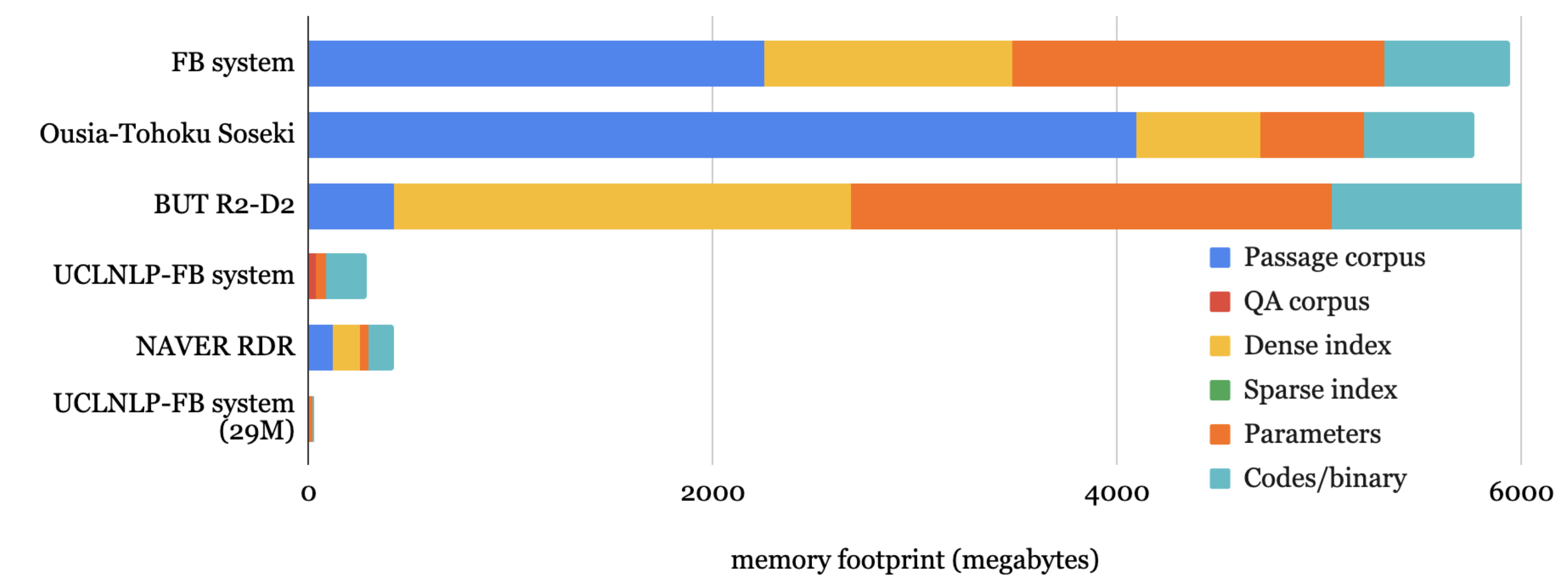}\vspace{-2em}
  \end{center}
  \caption{
  Memory footprint of each system component.}\label{fig:memory-footprint}
\end{figure}

We describe a set of provided baselines and systems from participants.
Systems from participants include the top 1--3 submissions per track based on automatic measure, considering margins between the submissions.
Table~\ref{tab:system-summary} summarizes the key distinctions between the systems, and Figure~\ref{fig:memory-footprint} shows the memory footprint of each system component.


\subsection{Unconstrained track}\label{subsec:unconstrained-track}
\ec{how many teams participate in this track?}
Both of the top two submissions make use of the previous state-of-the-art system, such as DPR~\citep{karpukhin2020dense} and a generative reader~\citep{izacard2020leveraging}. They are greatly enhanced by a better training objective, aggregating the answers from the extractive and generative models, or incorporating lists and tables from Wikipedia.
\paragraph{Baselines: REALM/DPR} Both REALM~\citep{guu2020realm} and DPR~\citep{karpukhin2020dense} use the retrieval-reader framework. They retrieve the top $K$ passages from Wikipedia by mapping the question and each passage to dense vectors and employing a maximum inner product search. The top $K$ passages are then fed into an extractive reader which predicts the start and the end position of the answer. A key distinction between REALM and DPR is that REALM is trained with a self-supervised, joint objective, while DPR is trained in a pipeline manner with gold positives and distantly supervised negatives.

\paragraph{\msunrestricted}
The UnitedQA system consists of three components: {\em Retrieval}, {\em Reading}, and  {\em Re-ranking}.
First, it uses DPR to fetch top 100 passages from the English Wikipedia dump for a given question. Second, a hybrid approach, combining both generative and extractive readers, is used to produce answer candidates from the collection of retrieved passages.
The generative reader is Fusion-in-Decoder~\citep{izacard2020leveraging} based on T5~\citep{t5}, and the extractive reader is based on ELECTRA~\citep{clark2020electra}.
Several techniques are applied to facilitate the training of the readers: posterior differential regularization (PDR, \citet{cheng2020posterior}, improved loss-term \citep{cheng-etal-2020-probabilistic} are explored for the extractive model, and adversarial training \citep{ju2019technical} approach, attention bias \citep{lewis2020pretraining} for the generative model. At the final re-ranking stage, it combines the generative and extractive model predictions with linear interpolation, and produce the final answer.
More details of the system can be found in~\cite{cheng2021unitedqa}.

\paragraph{\fbunrestricted}
This system uses a retriever-reader architecture.
The {\em retriever} is a combination of DPR and Generation Augmented Retrieval (GAR)~\citep{GAR}, where each of them retrieves 50 passages.
The DPR encoders are trained iteratively, where better hard negatives are mined at each step using the model at the previous step---two iterations are found to be sufficient.
Its dense index includes lists and tables, as well as regular text passages.
Specifically, 455907 tables and infoboxes are processed from the NQ training corpus using a trivial linearization, which concatanates the text representation of each row with a newline character.\footnote{This linearization is found to be more effective than converting each row to a statement.} Tables are chunked at 100 tokens with the header included in each chunk.
The {\em reader} is Fusion-in-Decoder based on T5-large, which is given 100 passages from the retriever and generates an answer.
More details of the system are in~\cite{oguz2020unified}.

\subsection{\medium{} track}\label{subsec:medium-track}
For this track, we provided two baselines. One retrieval-based, and one generative model.
However, all of the top three submissions are retrieval-based, and reduce the memory footprint by drastically pruning the Wikipedia corpus, either by a learned model or based on a page view.
They also made use of model quantization and the state-of-the-art in compression to limit their system's footprint. They additionally enhanced the previous state-of-the-art as done in an unrestricted track.

\paragraph{Retrieval-based baseline: DPR-subset}
We create a variant of DPR with pruned Wikipedia. Specifically, we only keep passages from the pages that are paired with the questions on the training set of NQ, reducing the number of passages from 21M to 1.6M.

\paragraph{Generative baseline: T5-XL+SSM} T5 \citep{t5} is a text-to-text Transformer language model utilizing an encoder-decoder architecture that was pre-trained using a ``span corruption'' objective on
Common Crawl.
This approach is based on \citet{roberts2020knowledge} which demonstrated a ``closed-book'' setting where the model was only able to access the knowledge stored in its parameters after fine-tuning on the task. Following \citet{roberts2020knowledge}, the model is additionally pre-trained using a ``salient span masking'' objective \citep{guu2020realm} before fine-tuning. The XL model has approximately 3B parameters.





\paragraph{\fbmedium}
This system is based on a retriever-reader approach.
The retriever is an ensemble of a dense retriever and GAR. The dense retriever is initialized from BERT-base and is trained by distilling the cross-attention scores of the reader.
GAR~\citep{GAR} is a retriever with generative query augmentation, based on BM25. The Lucene\footnote{\url{https://lucene.apache.org/}} index is built on the fly.
The reader is Fusion-in-Decoder initialized from T5 large. 

The text corpus initially included 26M passages including the plain text and the lists, but went through an article-level filtering to include 18.8M passages.
Specifically, it uses a linear classifier where each Wikipedia article is represented by its title and list of categories.

The model weights are stored using float16, taking 1.6GB of memory. The dense index is compressed by relying on three strategies, described in \cite{izacard2020memory}:
1) the vector representations go through dimension reduction from 768 to 256,
2) they are further quantized through product quantization, using 2 bits per dimension,
and 3) the text of Wikipedia is compressed using lrzip\footnote{\url{https://github.com/ckolivas/lrzip}}. 

\paragraph{\soseki}
Soseki is an open-domain QA system that adopts a two-stage approach consisting of \textit{Retriever} for passage retrieval, and \textit{Reader} for reading comprehension. Given a question, Retriever obtains top-k candidate passages from Wikipedia. Then, Reader selects the most relevant passage from them and extracts an answer from the passage.

Retriever is based on a re-implementation of DPR, where two BERT-base-uncased models are used to embed questions and passages. We also quantize passage embeddings to reduce the system size. Embeddings of Wikipedia passages are precomputed and stored using Faiss~\citep{JDH17}. Reader is a reading comprehension model based on ELECTRA-large~\citep{clark2020electra}. We added one dense layer for selecting the most relevant passage, and two dense layers for detecting the start and end positions of the answer. All modules are trained on the NQ dataset.

To reduce the system size under 6GB, we compressed the models, passages, and other data files using ZPAQ\footnote{\url{http://mattmahoney.net/dc/zpaq.html}} and excluded Wikipedia passages with less than 40 monthly page views, resulting in 18M passages.
These tricks do not drop the accuracy.

\paragraph{\brno}
R2-D2 is composed of a dense retriever, a re-ranker and two readers.
The dense retriever is based on RoBERTa~\citep{liu2019roberta} and is trained via an objective known from DPR.
It retrieves $K = 400$ passages from a pruned version of Wikipedia, reduced from 21M to 1.6M passages. Pruning is done via a simple binary classifier based on RoBERTa trained on the data created from the golden passages and negative passages randomly sampled from the index. This classifier obtains 90.2\% accuracy.
The re-ranker (based on Longformer~\citep{beltagy2020longformer}) concatenates retrieved passages, assigns a score for each passage, and selects $V = 20$ passages.
The extractive reader (based on ELECTRA) reads $V$ passages in a similar way as in \cite{fajcik2020rethinking}. It is trained via the marginal likelihood objective combined with the compound objective.
The generative reader follows the Fusion-in-Decoder schema and generates the answers.
R2-D2 aggregates the output from these two readers using two fusioning methods. First, it reranks the top spans from the extractive reader by feeding them to a generative reader and combines its log likelihood with the log likelihood from the extractive reader through a linear combination. They are then further aggregated with abstractive answers from the generative reader that are generated independently, through another linear combination.
The parameters are stored in float16 and compressed using ZIP.
More details of the system are in \citet{r2d2_2021}.


\subsection{\low{} track}\label{subsec:low-track}
The top performing submissions in this track made more drastic changes to get under \low.
They have completely different approaches from each other.

\paragraph{Baseline: T5-Small+SSM} This is a smaller version of the T5-XL+SSM baseline above, with 512 hidden dimension instead of 4096, 6 layers instead of 24, and 8 attention heads instead of 64. It contains 60M parameters in total.

\paragraph{\ucl}
This system is based on the approach from \cite{lewis2020question}, consisting of a database of question-answer pairs and a retriever which returns the answer of the most similar stored question to an input question.
This approach is attractive as it performs well using low parameter-count models, and less space is needed to store question-answer pairs than full Wikipedia passages.

The {\em Question-Answer Pair Generator} is similar to \cite{alberti-etal-2019-synthetic,lewis-etal-2019-unsupervised}.
First, a passage selection model $P(c)$ is trained to identify appropriate passages from Wikipedia.
For high-probability passages (w.r.t.\ $P(c)$), the system performs Named Entity Recognition to extract likely answers $a$, and generates questions $q$ from $(c,a)$ using a BART~\citep{DBLP:conf/acl/LewisLGGMLSZ20} model $P(q \mid a, c)$ fine-tuned on NQ.
Finally, the system filters question-answer pairs with a ``global consistency filter''---if an open-domain QA system~\citep{izacard2020leveraging} is given $q$ and generates an answer that is consistent with $a$, the $(q, a)$ pair is added to the database.
This is because generated $(q,a)$ pairs are sometimes incorrect or unanswerable without the passage they were generated from.
The final database consists of NQ, the EfficientQA development set and 2.3M generated question-answer pairs.
The {\em Retriever} consists of TF-IDF and a bi-encoder retriever based on ALBERT-base~\citep{lan2020albert}.
The {\em Reranker} is a cross-encoder module based on ALBERT-large;
it reranks the top-10 question-answer pairs from the retriever, and the answer of the top-1 question-answer pair is chosen as the final answer.

The system is further compressed using TFLite, quantization, and Alpine Linux.
More details can be found from \citet{lewis2021paq}.

\paragraph{\naver}
RDR is a lightweight retrieve-and-read system, consisting of a single Transformer (MobileBERT~\citep{mobilebert}) that performs both retrieval and reading, an index of dense passage vectors and the filtered Wikipedia corpus.
The Transformer serves as a dense retriever as DPR does with two differences: (1) a single Transformer is used to encode both the question and the passage, and (2) the embeddings are defined as the first 128 out of 512 dimensions of the output vectors from the Transformer that correspond to the {\tt [CLS]} token.
The same Transformer serves as an extractive reader by producing the scores of the start and the end of the answer span as well as the passage reranking score.
A distillation technique is used for training, by minimizing the KL divergence between its start, end, and reranking scores and those of a fully trained DPR reader.
The Transformer is further finetuned in an iterative manner as a retriever and as a reader. 

The index consists of 1.2M 128-dimensional INT8 vectors (scalar-quantized and unsigned), which are the dense embeddings of the subset of the 21M passages in Wikipedia, filtered by a RoBERTa~\citep{liu2019roberta}-based binary classifier trained with logistic regression to exclude uninformative passages.
Positives to train this classifier are the top 200 passages for each question on NQ dataset and EfficientQA development set, retrieved by~\cite{yang2020retriever}, a DPR retriever further finetuned on hard negatives using knowledge distillation from a DPR reader.
Negatives are uniformly drawn from the set of 21M passages, excluding positives.
More details can be found from \citet{yang2021designing}.

\subsection{\smallacc{} smallest track}\label{subsec:smallest-track}

\paragraph{Baseline: T5-XL+SSM} Among the provided baselines, the smallest system with an accuracy of over \smallacc{} is T5-XL+SSM, the same system described in Section~\ref{subsec:medium-track}. This system is 5.65GB and achieves an accuracy of 28\%.


\paragraph{\uclsmall}
This system is the same system as \ucl\ in the \low\ track with following minor modifications, to further decrease the memory: (1) the retriever is just TF-IDF instead of a combination of TF-IDF and a bi-encoder, (2) the reranker uses ALBERT-base instead of ALBERT-large, and (3) there are 40k generated question-answer pairs, instead of 2.3M.

\section{Human Annotations of Correctness}\label{sec:annotation}

%
%
%
%
%
The EfficientQA development and test sets have up to five reference answers per question.
Due to the variability of language, these five answer strings are often not exhaustive, and systems predict correct answers that are judged incorrect according to the automatic metrics. 
To rectify this, we sent predictions from each of the systems in Section~\ref{sec:systems} for human rating to get a better estimation of accuracy.

Each system prediction was sent for rating by three separate annotators:
1) the annotator first works on understanding the meaning and intent of the question (with a web search if necessary).
2) The annotator then determines whether the question is ambiguous, i.e., whether the question can lead to multiple different answers depending on factors such as: when the query was asked; where the query was asked; some unspoken intent of the questioner; or the opinion of the person giving the answer.
3) Finally, the annotator determines whether each answer is ``definitely correct'' (correct given a usual interpretation of the question), ``possibly correct'' (could be correct, given some interpretation of the question), or ``definitely incorrect''. 

Since the original NQ data was collected more than a year before the start of the EfficientQA competition, the denotation of some questions may have changed over time (e.g. ``who won the last season of bake-off'').
Rather than determine a single correct point in time for these questions, we asked our annotators to assume that the query could have been asked at any time since the web has existed \jbg{when was this, to be precise?} and choose the ``possibly correct'' label for answers that may or may not have been correct when the question was asked.

The final rating is an aggregation of ratings from three annotators: if at least 2/3 raters determined it to be ``definitely correct'', the label is ``definitely correct''. If at least 2/3 raters determined it to be either ``definitely correct'' or ``possibly correct'', the label is ``possibly correct''.
The pairwise agreements of the human ratings are 69.2\% (Cohen's $\kappa=53.8$) for 3-way ratings, 85.7\% (Cohen's $\kappa=71.4$) for whether the prediction is definitely correct or not, and 76.7\% (Cohen's $\kappa=53.4$) for whether the prediction is possibly correct or not.\footnote{An agreement in whether the question is ambiguous or not is 61.3\% (Cohen's $\kappa=22.6$).
We conjecture the low agreement is due to an intrinsic difficulty in finding ambiguity~\citep{min2020ambigqa}, and consider the question to be ambiguous if at least one annotator rated it as ambiguous.
}



\begin{table}[t]
\begin{center} \footnotesize
\begin{tabular}{l|l|cc>{\color{gray}}c}
\toprule
    \multirow{2}{*}{Track} & \multirow{2}{*}{Model} & \multirow{2}{*}{Automatic eval} & \multicolumn{2}{c}{Human eval} \\
    &&& Definitely & \color{black} Possibly \\
\midrule
    \multirow{2}{*}{Unrestricted} & 
    \msunrestricted & 54.00 & 65.80 (+21.9\%) & 78.12 (+44.7\%) \\
    & \fbunrestricted & 53.89 & 67.38 (+25.0\%) & 79.88 (+48.2\%) \\
\midrule
    \multirow{3}{*}{\medium} &
    \fbmedium       & 53.33 & 65.18 (+22.2\%) & 76.09 (+42.7\%) \\
    & \soseki         & 50.17 & 62.01 (+23.6\%) & 73.83 (+47.2\%) \\
    & \brno           & 47.28 & 58.96 (+24.7\%) & 70.33 (+49.2\%) \\
\midrule
    \multirow{3}{*}{\low} &
    \ucl            & 33.44 & 39.40 (+17.8\%) & 47.37 (+41.7\%) \\
    & \naver          & 32.06 & 42.23 (+31.7\%) & 54.95 (+71.4\%) \\
\midrule
    {\scriptsize \smallacc\ smallest} &
    \uclsmall       & 26.78 & 32.45 (+21.2\%) & 41.21 (+53.9\%) \\
\bottomrule
\end{tabular}
\caption{\small Summary of the result. For human evaluation result, relative improvements over the automatic evaluation are indicated in parenthesis. Following our analysis of the annotations, we use `Definitely correct' human ratings as a primary metric.}\vspace{-3.5em}
\label{tab:result-summary}
\end{center}
\end{table}

\section{Results \& Analyses}\label{sec:results-and-analysis}

\subsection{Results}\label{sec:results}

All of the five systems in the unrestricted track and the \medium\ track significantly outperform the state-of-the-art (Table~\ref{tab:result-summary}) at the beginning of the competition---DPR (36.6\%) and REALM (35.9\%). Systems in the \medium\ track approach the unrestricted track's accuracy; for instance, the accuracy \fbmedium\ is comparable to the accuracy of the top systems in the unrestricted track.
The improvements in the \low\ track are also impressive; both the top two systems significantly beat T5-small (17.6\%).

\paragraph{Discrepancy between automatic eval and human eval}
Human raters find 13\% and 17\% of the predictions that do not match the reference answers to be definitely correct or possibly correct, respectively, overall increasing the accuracy of the systems.
Most systems showed 17--25\% and 41--54\% improvement in accuracy when using definitely correct and possibly correct human evaluation respectively, compared to automatic evaluation metric which only consider exact string match to existing reference answers. An exception is \naver{}, which achieves significantly larger improvements (32\% and 71\%, respectively).
We also found that when the gap in automatic measure between systems is marginal (around or smaller than 1\%), human evaluation may change the rankings between the models. 

\ifsqueeze
\else
  \begin{figure} 
  \begin{center}
    \includegraphics[width=0.8\textwidth]{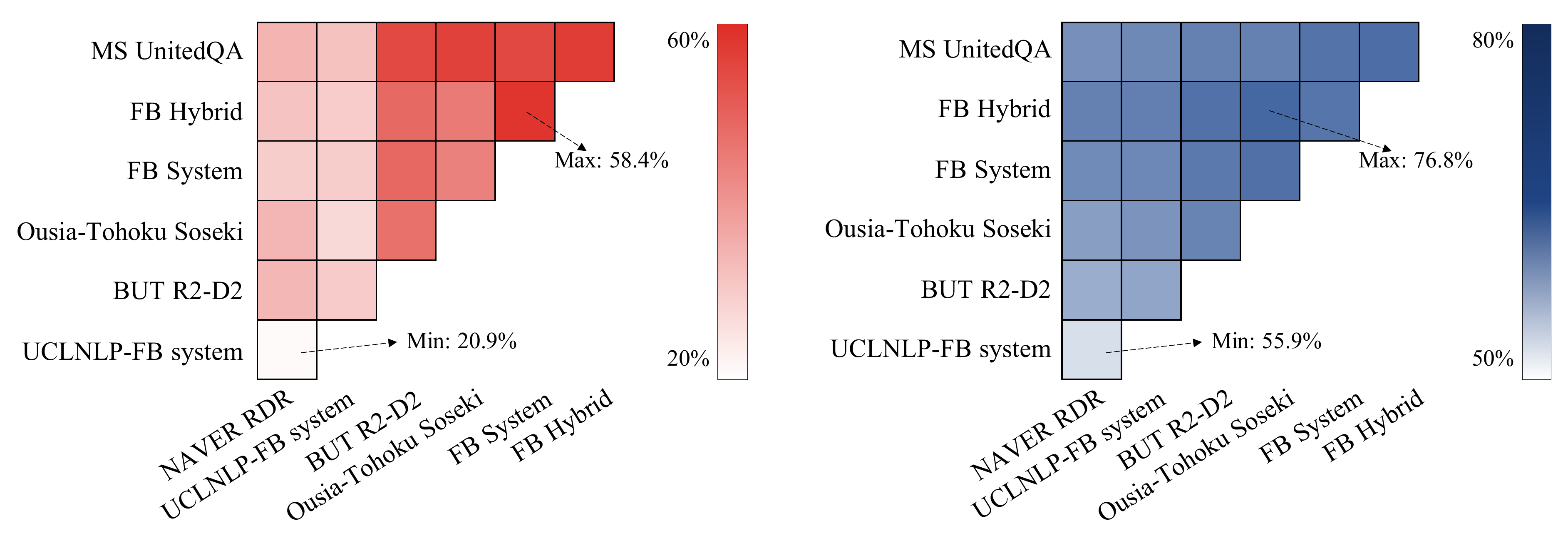}\vspace{-2em}
  \end{center}
  \caption{
  {\em (Left)} Agreement between system predictions.
  {\em (Right)} Ensemble oracle accuracy, which considers a prediction correct if at least one of the system predictions is correct (based on ``definitely correct'' human evaluation).}\label{fig:system-agreement}
\end{figure}

\paragraph{Agreement between system predictions}
Figure~\ref{fig:system-agreement} (left) shows the agreement between system predictions, based on exact match in automatic evaluation. The largest agreement is made between \fbunrestricted{} and \fbmedium{}, likely because they are both based on DPR and Fusion-in-Decoder.
Agreements between systems in the unrestricted and the \medium{} tracks are generally higher, likely because they are all based on retrieval-reader framework and pruning Wikipedia does not hurt too much.
Two systems in the \low{} track have smaller agreement with the other systems, and agree with each other even less.

\paragraph{Ensemble oracle accuracy of the systems}
Figure~\ref{fig:system-agreement} (right) reports the ensemble oracle accuracy for each system pair, which considers a prediction to be correct if either system prediction is correct.
The \fbunrestricted{} \& \soseki{} pair achieves the highest ensemble oracle accuracy, indicating that their system predictions are substantially different from each other compared to other pairs with top performing systems.
\fi

\subsection{Analyses}\label{sec:analysis}

\jbg{Could we have a more informative title?}

We present an analysis of the 50 sample questions where at least one prediction does not match with the gold answer, but is judged as correct by human raters, with a ``definitely correct'' label or a ``possibly correct'' label, respectively.
The samples are largely divided into three classes: \textbf{as valid as gold} (judged as either definitely correct or possibly correct), \textbf{valid but not the best} (judged as possibly correct), and \textbf{closer to invalid} (judged as possibly correct).
We describe fine-grained categories and their percentage\footnote{
    The total is over 100\%; one question may have multiple predictions that fall into different categories.
}(\red{definitely correct} and \blue{possibly correct}) here, with
examples shown in Appendix~\ref{app:analyses} (Table~\ref{tab:analysis-examples}).

\vspace{.5em}
The following describes categories on predictions that are as valid as the gold answers. \vspace{-.1em}
\begin{itemize}\setlength\itemsep{0em}
  \item \textbf{Semantically the same} \stat{60}{22}: The prediction is semantically equivalent to the gold, e.g., ``about 930 BCE'' and ``around 930 BCE''.\footnote{A few cases in this category are due to inconsistent tokenization (e.g., ``Mercedes-Benz'' and ``Mercedes - Benz''), which could potentially be automatically captured but is not handled by the automatic evaluation from the literature.}
  \item \textbf{Open-ended question} \stat{6}{4}: There is a large set of distinct, plausible answers, mainly because the question is vague or open-ended.
  \item \textbf{Ambiguous entity/event references} \stat{20}{20}: There is a set of distinct answers because the question contains ambiguous references of entities or events. For instance, ``Gold woman'' in the example in Table~\ref{tab:analysis-examples} may refer to the fictional character (``Ayesh'') or the actress (``Elizabeth Debick'').
  \item \textbf{Different granularity} \stat{14}{6}: The prediction and the gold answer have different granularity, such as the year (1982) vs. the month (October 1982), or the city (Pawtucket) vs. the state (Rhode Island). 
  \item \textbf{Incorrect gold} \stat{2}{6}: The gold annotation is incorrect. 
 \end{itemize}
 
The following describes categories on possibly correct predictions that are as valid but not the best answers. \vspace{-.1em}
\begin{itemize}\setlength\itemsep{0em}
  \item \textbf{Ambiguous answer type} (\blue{20\%}): There is a unique entity or event that can be the answer to the question, but there is ambiguity on which exact text should be presented as the answer. For instance, both the episode title (``Somber News'') and the air date (``February 23 2013'') are the valid answer to the question in Table~\ref{tab:analysis-examples}.
  \item \textbf{Answer is time-dependent} (\blue{18\%}): The answer depends on the time of the question being asked. Questions in this category usually involve recurring events such as sports games or elections.
\end{itemize}

Finally, we present categories for predictions that are closer to invalid answers. \vspace{-.1em}
\begin{itemize}\setlength\itemsep{0em}
  \item \textbf{Conflicting information in Wikipedia} (\blue{4\%}):
  While the gold is the only valid answer, the English Wikipedia contains incorrect information that supports that prediction is the answer to the question. Consider the question in Table~\ref{tab:analysis-examples}. While ``Fort Hood'' is an incorrect answer by a
  fact, the Wikipedia page\footnote{ \url{https://en.wikipedia.org/wiki/Fort_Hood}} states ``Fort Hood is the most populous U.S. military installation in the world.''\footnote{At the human vs. computer competition, the human team also predicted ``Fort Hood''.}
  \item \textbf{Plausible only in certain conditions} (\blue{4\%}): Prediction may be valid in certain conditions, but is incorrect in general cases. For instance, for the question in the table, ``president of India'' may only be valid in India.
  \item \textbf{Mismatch with question intent} 
  (\blue{8\%}): The prediction is somewhat valid but supposedly not the intended answer to the question. For instance, the example question in Table~\ref{tab:analysis-examples} is supposedly asking for the answer that is different from the great depression.
  \item \textbf{Incorrect prediction} (\blue{2\%}): The prediction is definitely incorrect. 
\end{itemize}

Our two main takeaways are as follows.
First, the automatic evaluation confirms the observation of \cite{voorhees-00} that it is insufficient in capturing semantically equivalent answers, which are responsible for 60\% of the definitely correct predictions.
Second, ambiguity arises frequently in the questions in different levels, allowing multiple, semantically different answers to be valid.
This is a consistent with \cite{min2020ambigqa}, which reported that around half of the questions in NQ contain ambiguity, due to ambiguous references of entities, events or properties, or time-dependency of the answer.
Based on our human evaluation, annotations on ambiguity have low agreement rate (61.3\%, Cohen's $\kappa=22.6$), and predictions with the same level of plausibility are often marked as ``definitely correct'' or ``possibly correct'' by different human raters.
We note the notions of ``correctness'' and ``plausibility'' are not binary, and are instead often dependent on pragmatic interpretation of the questioner's intent. 
For example, the question ``who has the most superbowl rings'' could be read as ``which person (including coaches) has the most superbowl rings'', ``which player has the most superbowl rings'', ``which team has the most superbowl rings''.
All three annotators identified this question as being ambiguous but they disagreed about the validity of the different readings. The three raters were split three ways when rating the correct answer (``Pittsburgh Steelers'') for the last interpretation.
Meanwhile there were no ``incorrect'' ratings, and 2/3 ``definitely correct'' ratings given to the correct answer (``Tom Brady'') for the second interpretation, despite the fact that two coaches have more superbowl rings.
Clearly, the annotators are applying some personal interpretation of the questioner's intent and answer plausibility.

While we believe that many real world questions do require some non-literal assumptions about the questioner's intent, and we believe that the natural language processing community should not shy away from that task, we also acknowledge that there is work to be done in creating better, non-binary, definitions of correctness.
Section~\ref{sec:trivia} contrasts these interpretations of correctness with the more rigid definition used by the Trivia community.

\begin{table}[t]
\begin{center} \footnotesize
\begin{tabular}{l|cccc}
\toprule
    \multirow{2}{*}{Model} & \multicolumn{2}{c}{All} & \multicolumn{2}{c}{Unambiguous Qs} \\
    & Definitely & Possibly & Definitely & Possibly \\
\midrule
    \msunrestricted &   65.80&	78.12&	78.24&	81.18   \\
    \fbunrestricted &   67.38&	79.88&	82.65&	85.59   \\
    \fbmedium       &   65.18&	76.09&	79.12&	81.47   \\
    \soseki         &   62.01&	73.83&	72.94&	75.00   \\
    \brno           &   58.96&	70.55&	69.71&	72.06   \\
    \ucl            &   39.40&	47.37&	42.06&	43.24   \\
    \naver          &   42.23&	54.95&	49.71&	53.24   \\
    \uclsmall       &   32.45&	41.21&	28.53&	30.29   \\
\bottomrule
\end{tabular}
\caption{\small Human evaluation on the original set and a subset of unambiguous questions.}\vspace{-1em}
\label{tab:result-summary-unambiguous}
\end{center}
\end{table}

\paragraph{Performance on unambiguous questions}
To better understand the effect of ambiguity on the ranking of different solutions, we also evaluate system performance on a subset of the questions that are unambiguous. 
We define unambiguous questions to be those that (1) have at least three out of five reference answers contain valid short answers\footnote{This follows the original NQ approach of using annotator agreement to set a threshold for high quality answers.}, and (2) are not labeled as ambiguous by any of three human raters, resulting in 51.5\% of the original set.
Table~\ref{tab:result-summary-unambiguous} shows human evaluation on the original set and this subset of unambiguous questions.
Most systems, except \ucl, achieve higher accuracy on unambiguous questions, with the first three systems achieving over or near 80\%. Unsurprisingly, the gap between ``definitely correct'' accuracy and ``possibly correct'' accuracy is marginal on unambiguous questions.


Importantly, the overall rankings are unchanged when we restrict our evaluation set to only unambiguous answers.
This suggests that, while question ambiguity may lead to disagreement between annotators at a per-example level, it is not adversely impacting our ability to consistently rank solutions. 
%
More analyses can be found in Appendix~\ref{app:analyses}.


\ifsqueeze
\else
  \begin{figure}[htp!]
  \begin{center}
    \includegraphics[width=0.48\linewidth]{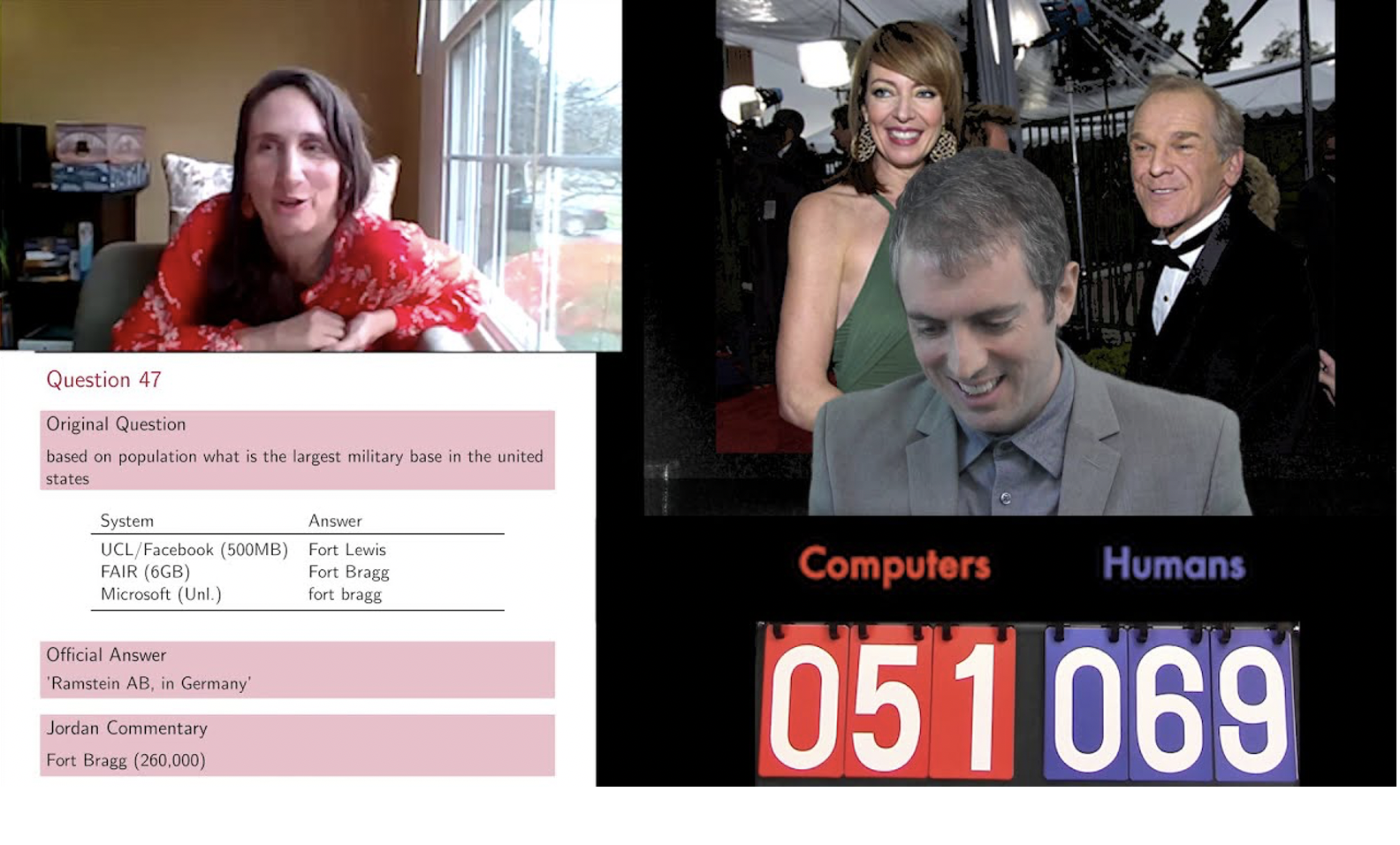}
    \includegraphics[width=0.5\linewidth]{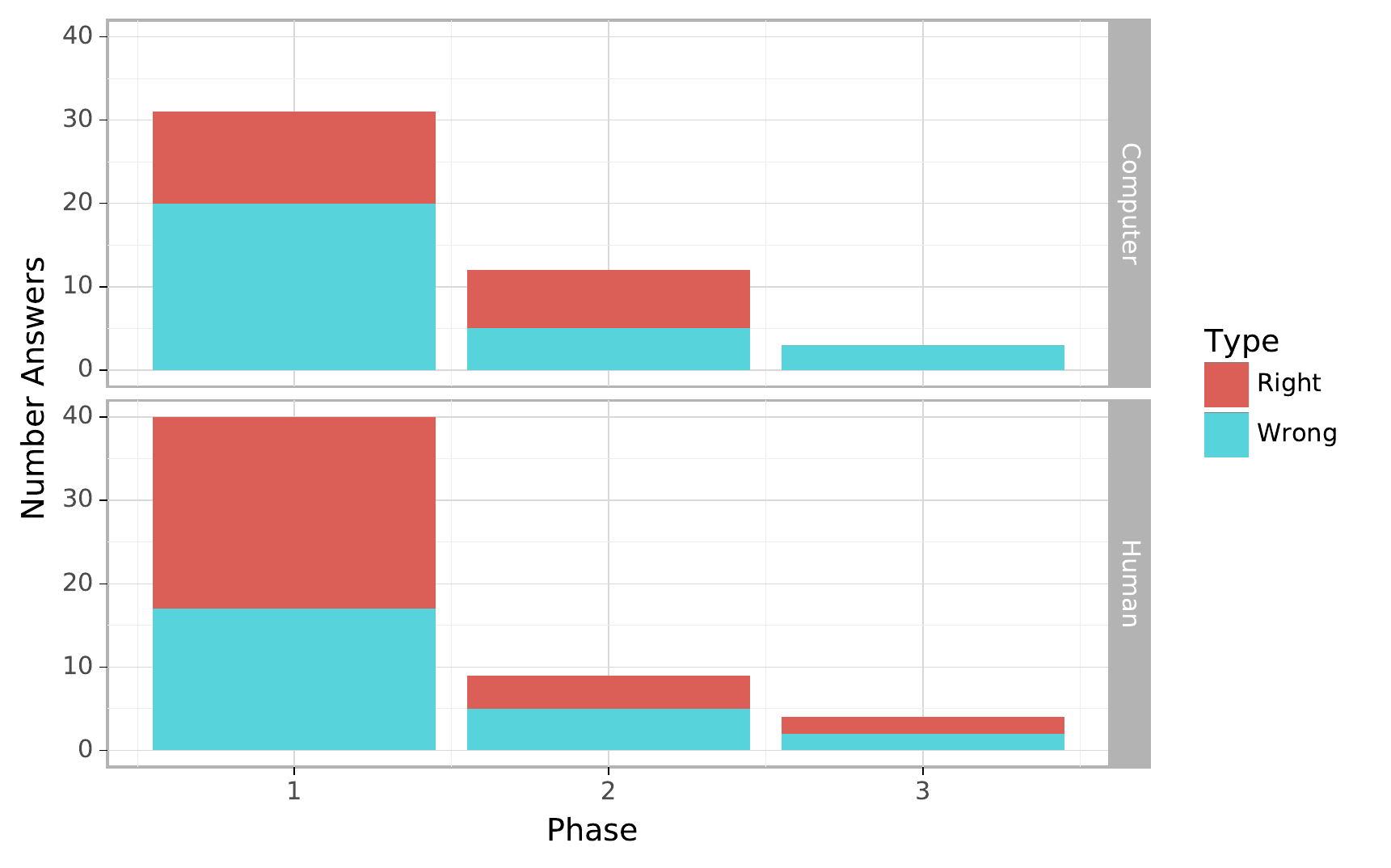}
  \end{center}\vspace{-2em}
  \caption{\small
  {\em (Left)} Screenshot from our final game between computer systems and trivia experts.  Full videos of the human--computer competition are available at  \url{https://go.umd.edu/2020eqa}.
  {\em (Right)} The human--computer competition is broken into three phases, where each phase allows the participants to use more resources.  While most questions were answered in Phase~1 (where humans had a distinct advantage), in Phase~2 computers had a slight advantage; 
  there were diminishing returns in Phase~3.
  }\label{fig:hcqa-accuracy-by-phase} \vspace{-2em}
\end{figure}

\section{Trivia Experts vs Computer Systems}\label{sec:trivia}

The questions in NQ are posed by humans to computers, and the competition attracted some of the strongest and most efficient QA systems available today.
However, humans also answer questions for fun and recreation~\citep{jennings-06} and the ultimate goal of artificial intelligence is to create machines that answer questions as well as humans~\cite[known as the Turing test]{Turing-95}.
Moreover, existing comparisons of human question answering ability often use unskilled humans~\citep{rajpurkar-16}, leading to claims of computers ``putting millions of jobs at risk''~\citep{cuthbertson-18}.
Or, in competitions with trivia experts~\citep{ferruci-10}, arcane rules of competitions can tilt the playing field toward computers~\citep{boyd-graber-20} or use unnatural questions~\citep{boyd-graber-12,DBLP:journals/corr/abs-1904-04792}.

\subsection{A Fair Comparison}

We advertised our competition to trivia enthusiasts on social media.
Teams of up to eight players applied to be part of the competition.
We selected five teams to participate in the preliminary competition (results in Section~\ref{sec:hcqa-prelims}).

To create a fair competition and to showcases all of the tiers of the efficient QA competition, we offered three ways to answer each question where either humans or computers have more resources to answer a question (Table~\ref{tab:hcqa-phases}).

\begin{table*}[t]
    \parbox{.6\linewidth}{\centering \footnotesize
    \setlength\tabcolsep{3pt}
    \begin{tabular}{lccr}
         {\bf Phase} & {\bf Human} & {\bf Computer} & {\bf Points} \\
         \toprule
         1 & Single player ``buzz in'' & \low{} & 3 \\
         2 & Team discussion & \medium{} & 2 \\
         3 & Team discussion w/ search results & Unlimited & 1 \\
         \bottomrule
    \end{tabular}
    \caption{Phases of the human--computer competition.  On odd questions the humans go first, while on even questions the computers go first.  If neither team has a correct answer in a phase, we move on to the next phase.}
    \label{tab:hcqa-phases}
    }
    \hfill
    \parbox{.35\linewidth}{\centering \footnotesize
    \begin{tabular}{cc}
    {\bf Team} & {\bf Margin} \\
    \toprule
        B & 0.4 ppq \\
        C & 0.1 ppq \\
        D & -0.1 ppq \\
        A & -0.2 ppq \\
        E & -0.6 ppq \\
        \bottomrule
    \end{tabular}
    \caption{Marin per question of the human teams at the preliminary competition.
    }
    \label{tab:hcqa-points}
    }
\end{table*}
To complement the \low{} systems, humans had to instantly signal when they knew the answer to a question.  
This reflects instant recall of a fact by a single individual.
In the next phase in competition with the \medium{} systems, both humans and computers had more resources: the human team could discuss the answer for thirty seconds, arguing why they believe their answer is correct and computers had over ten times the memory.
Finally, to focus one reading comprehension, unlimited systems faced off against the human teams who also had access to snippets from search results using the question as a query.
As with the previous phase, they have thirty seconds to discuss their answer.

We selected questions for the human eval based on the following criteria:
\begin{itemize}\setlength\itemsep{0em}
    \item Diverse over topic, ensuring there were questions about history, literature, philosophy, sports, and popular culture.  This results in fewer questions about sports and popular culture than the standard NQ distribution.
    \item Not tied to 2018; 
    we excluded questions that depend on the date being asked. 
    \item Interesting questions.  While not strictly adversarial, we wanted to showcase both human and computer ability, so we excluded questions that many humans would not know (e.g., ``how many us states are there'') or questions with answers that are difficult to evaluate in the NQ framework (``how many words are in \textit{Les Mis\'erables}?'').
    \item To avoid the issues described in Section~\ref{sec:annotation}, we avoided questions that were overly ambiguous 
    (answer changes based on time the question was asked, unclear answer type, mismatch with question intention, etc).  
\end{itemize}

Thus, for the human competition, we exclude the broader interpretation of correct adopted in Section~\ref{sec:analysis} to a standard we called ``actually correct'', consistent with traditional trivia interpretations~\cite{jennings-06}.
A human judge researched all of the answers to the questions and evaluated whether an answer was correct or not.
This allows for more permissive answers lines such as ``to pass away'' to be accepted for questions like ``what does it mean to cross over the rainbow bridge'' even though the answer line only lists ``heaven''.
Again, consistent with trivia best practice, we only required full names in the case of confusion (e.g., ``Theodore Roosevelt'' or ``Franklin D. Roosevelt'' instead of just ``Roosevelt'').

\subsection{Preliminary Competition}
\label{sec:hcqa-prelims}

To select which of the human teams faced off against the top computers in the final competition and to let the humans teams practice this unconventional format, we had the human teams face off against the baseline systems: T5 (\low{}), DPR (\medium{}), and REALM (Unlimited).

We set aside an hour for each team.  Because better teams can get through more questions, to normalize comparisons per team, we computed the average margin per question (Table~\ref{tab:hcqa-points}).
Only two teams had higher scores than the computer baseline teams.
The top team (Team~B), which had multiple professors and trivia champions was clearly the best human team on this set of questions.


\subsection{Final Game}


The winning team (Team B) went up against the winning computer systems in a final match with fifty questions.
Most questions were answered in Phase~1, and those that were not were typically answered in Phase~2 (Figure~\ref{fig:hcqa-accuracy-by-phase} (right)).
Only four questions reached Phase~3; these questions that stumped both humans and computers:\vspace{-.1em}
\begin{itemize}\setlength\itemsep{0em}
    \item Kimi no na wa what does it mean
    \item Who said there are old pilots and there are bold pilots
    \item What is the second largest city in ethiopia
    \item Who said i'll make mincemeat out of you
\end{itemize}
Of these questions, only the second largest city in Ethiopia was answered correctly (the humans were working through a list of cities in Ethiopia they knew).
The other questions represent questions that cannot be answered by Wikipedia (``old pilots''), questions that require language capabilities beyond question answering (``kimi no na wa''), or that have a more nuanced answer than NQ can provide.
For example, on the questions ``who said i'll make mincemeat out of you'', the human team reasonably thought that the question meant who \emph{originated} the phrase, while NQ's answer was the cartoon character who is most associated with the phrase (\underline{Klondike Kat}).

\subsection{Reflection}
\label{sec:reflection}

For question answering, there are multiple interpretations of the phrase ``human evaluation''.  
Human evaluation of answers are important for exposing problems in the dataset, revealing ambiguity, and measuring whether the answers are useful.
However, for the ultimate questions of where we have achieved artificial intelligence~\citep{Turing-95}, we need fair comparisons with \emph{skilled} humans.  
Moreover, to measure whether question answering systems are \emph{useful} for people, we need to create socio-technical systems where humans and computers can answer questions together~\citep{feng-19}.
%
More importantly,
trivia games are fun.  They help illustrate the strengths and weaknesses of the underlying dataset and the QA methods with a spoonful of sugar to make otherwise dull evaluations more interesting.

\fi


\section{Conclusions and Future Work}\label{sec:conclusion}

\jbg{This seems to repeat a lot of what's been said before.  Perhaps we can trim here or add new observations?}
\sm{Great point; maybe good to add discussions that are not too similar to discussions already mentioned? (or related to broader open-domain QA?)}

The EfficientQA competition was held to encourage research in open-domain question answering that focuses on both accuracy and memory footprint.
All top performing submissions have a retriever-reader framework.
Submissions to the unrestricted and the \medium{} track are enhanced over the state-of-the-art that retrieves a subset of Wikipedia passages and employs the extractive and/or generative answer modules; 
systems in the \medium{} track additionally cleverly select a small subset of Wikipedia, only marginally sacrificing accuracy when combined with state-of-the-art compression.
In more restricted tracks, systems explore novel approaches and achieve impressive improvements over the baselines.
They either generate a large corpus of question-answer pairs and retrieves the closest question to the input question, or dramatically filter Wikipedia and use a single Transformer model for retrieval and answer extraction.
Still, they are behind the unrestricted systems in performance by 20\%, indicating significant room for improvements in memory-restricted settings.

A human analysis shows that automatic evaluations of QA systems are not sufficient for thorough evaluations.
Human raters find 30\% of the predictions that do not match reference answers but are nonetheless correct.
This does not affect all systems equally: relative accuracy rises using definitely correct (between 18--32\%) and possibly correct (between 42 and a whopping 71\%).
The rise is mainly due to automatic evaluation failing to capture semantically equivalent answers,  time-dependence of the answers, or underlying ambiguity in the questions~\citep{min2020ambigqa}.


Future work in efficient open-domain question answering should continue to explore the tradeoff between system size, accuracy, and abstaining~\citep{he-16,rajpurkar-18}.
Moreover, it is important to continue to refine the quality of QA evaluation: not all annotation and not all annotators are created equal.
Using trivia enthusiasts for annotation and human benchmarks is a fun and effective evaluation of relative computer QA ability.
We would encourage future leaderboards to use some element of human verification in important evaluations (e.g., an annual bake off), removing ambiguity in questions~\citep{min2020ambigqa}, crafting adversarial examples~\citep{jia-17,wallace-19,dua-19,bartolo-20}, or evaluating whether a response is useful to a user~\citep{fan2019eli5, feng-19}.
\ec{can we add a paragraph summarizing what resources will be released from the competition?}

\acks{
We thank all the participants for taking part and making this a successful competition.
We thank Google for providing prizes for computer participants.
Boyd-Graber is 
supported by {\sc nsf} Grant {\sc iis}-1822494.
Any opinions, findings,
conclusions, or recommendations expressed here are those of the
authors and do not necessarily reflect the view of the sponsor.
}

\bibliography{bib/journal-full,bib/nqcomp,bib/jbg}

\clearpage
\appendix

\section{Analyses}\label{app:analyses}

\paragraph{Characteristics of {\em easy} and {\em hard} questions}

Table~\ref{tab:analysis-easy-hard} shows questions that are answered correctly by all seven systems ({\em easy}), or answered incorrectly by all seven systems ({\em hard}).
A common feature of easy questions~\citep{sugawara-18,kaushik-18} is that a sentence from Wikipedia provides an explicit support to the question. Such supporting sentences have high lexical overlap with the question and require little paraphrasing (e.g. ``the highest population'' vs. ``the most populous''). Even when the question is not well-formed, all systems may find the correct answer if there is high lexical overlap between the question and the supporting sentence (e.g. the fifth example question in Table~\ref{tab:analysis-easy-hard}).

Many of the hard questions have their answers in the tables (e.g. the first question in Table~\ref{tab:analysis-easy-hard}). This likely makes them ``hard'' because all systems except \fbunrestricted{} do not consider the tables, and even \fbunrestricted{} may miss such cases.
In the other cases, systems make mistakes even when there is the text that supports the answer (e.g. the next two questions in Table~\ref{tab:analysis-easy-hard}). Though the reason for the mistake would be different depending on the systems, we conjecture that such supporting sentences are harder to retrieve or do not support the answer as explicitly as other examples. Occasionally, the question is not well-formed, such as the last one in Table~\ref{tab:analysis-easy-hard}.

\begin{table}[t]
\begin{center}\small
\begin{tabular}{lcc}
    \toprule
        Category &  definitely (\%) & possibly (\%)  \\
    \toprule
        \textbf{\em As valid as gold} \\
        Semantically the same               & 60    & 22 \\
        Open-ended question                 & 6     & 4 \\
        Ambiguous entity/event references   & 20    & 20 \\
        Answer has different granularity    & 14    & 6 \\
        Gold is incorrect                   & 2     & 6 \\
    \midrule
        \textbf{\em Valid but not the best} \\
        Ambiguous answer type               & 0     & 20 \\
        Answer is time-dependent            & 0     & 18 \\
    \midrule
        \textbf{\em Less plausible} \\
        Conflicting info in Wikipedia       & 0     & 4 \\
        Plausible only in certain condition & 0     & 4 \\
        Mismatch with question intent    & 0     & 8 \\
        Incorrect prediction                & 0     & 2 \\
    \bottomrule
\end{tabular}
\caption{
    Analysis of predictions that are rated as {\em definitely correct} or {\em possibly correct} by human raters.
    Examples of each category are shown in Table~\ref{tab:analysis-examples}.
    Note that the total is over 100\% as one question may fall into multiple categories.
}
\label{tab:analysis-classification}
\end{center}
\end{table}

\begin{table}[t]
\begin{center} \setlength\tabcolsep{5pt} \footnotesize 
\begin{tabular}{ll}
    \toprule
        Category & Example \\
    \toprule
        \textbf{\em As valid as gold} \\
    \midrule
        \multirow{4}{*}{Semantically the same}
            & Q: When did israel split into israel and judah \\
            & Gold: about 930 BCE / Prediction: around 930 BCE \\
            & Q: What movie is bring it on the musical based off of \\
            & Gold: Bring It On / Prediction: 2000 film of the same name \\
    \midrule
        \multirow{2}{*}{Open-ended question}
            & Q: City belonging to mid west of united states \\
            & Gold: Des Moines / Prediction: kansas city\\
    \midrule
        \multirow{2}{*}{Ambiguous entity/event references}
            & Q: Gold woman in guardians of the galaxy 2 \\
            & Gold: Ayesha / Prediction: Elizabeth Debicki \\
    \midrule
        \multirow{4}{*}{Answer has different granularity}
            & Q: When did the musical cats open on broadway \\
            & Gold: 1982, 7 October 1982 / Prediction: october 1982 \\
            & Q: Where did the industrial revolution in the united states begin \\
            & Gold: in Pawtucket / Prediction: rhode island \\
    \midrule
        \multirow{2}{*}{Gold is incorrect}
            & Q: When did universal studios become a theme park \\
            & Gold: 1965 / Prediction: july 15 1964, 1964 \\
    \toprule
    \textbf{\em Valid but not the best} \\
    \midrule
        \multirow{2}{*}{Ambiguous answer type}
            & Q: When did naruto find out about jiraiya's death \\
            & Gold: ``Somber News'' / Prediction: february 23 2013 \\
    \midrule
        \multirow{2}{*}{Answer is time-dependent}
            & Q: Who goes to the big 12 championship game \\
            & Gold: Oklahoma Sooners / Prediction: tcu horned frogs \\
    \toprule
    \textbf{\em Less plausible} \\
    \midrule
        \multirow{2}{*}{Conflicting info in Wikipedia}
            & Q: Based on population what is the largest military base in the \\
            & united states / Gold: Fort Bragg / Prediction: fort hood \\
    \midrule
        \multirow{2}{*}{Plausible only in certain condition}
            & Q: Who appointed the chief justice of high court \\
            & Gold: President / Prediction: president of india \\
    \midrule
        \multirow{2}{*}{Mismatch with question intent}
            & Q: While the united states dealt with the depression it was also facing \\
            & Gold: Racial tensions / Prediction: great depression, economic downturns \\
    \midrule
        \multirow{2}{*}{Incorrect}
            & Q: When were banded iron formations formed on the sea floor \\
            & Gold: Precambrian / Prediction: some 185 billion years ago \\
    \bottomrule
\end{tabular}
\caption{
    Examples of predictions rated as {\em definitely correct} or {\em possibly correct} by human.
}
\label{tab:analysis-examples}
\end{center}
\end{table}

\begin{table}[t]
\begin{center}\small
\begin{tabular}{l}
    \toprule
    \textbf{\em \textcolor{blue}{Easy} questions} \\
        Q: What city in america has the highest population / Gold: New York City \\
        {\em New York City}: New York City is the most populous city in the United States ... \\
    \midrule
        Q: Who plays 11 in stranger things on netflix / Gold: Millie Bobby Brown \\
        {\em Millie Bobby Brown}: She gained notability for her role as Eleven in the first season of Netflix science \\ fiction series Stranger Things. \\
    \midrule
        Q: Where did leave it to beaver take place / Gold: Mayfield \\
        {\em Leave It to Beaver}: Leave It to Beaver is set in the fictitious community of Mayfield and its environs. \\
    \midrule
        Q: The substance that is dissolved in the solution / Gold: solute \\
        {\em Solution}: A solute is a substance dissolved in another substance, known as a solvent. \\
    \midrule
        Q: This means that in dna cytosine is always paired with / Gold: guanine \\
        {\em Guanine}: In DNA, guanine is paired with cytosine. \\
    \toprule
    \textbf{\em \textcolor{red}{Hard} questions} \\
        Q: What is the area code for colombo sri lanka / Gold: 036 \\
        Telephone numbers in Sri Lanka: (Answer in the table) \\
    \midrule
        Q: Which country is highest oil producer in the world / Gold: United States / Predictions: Russia \\
        {\em History of the petroleum industry in the United States}: For much of the 19th and 20th centuries, the \\ US was the largest oil producing country in the world. \\
        {\em  List of countries by oil production}: (Answer in the table) \\
        {\em Russia}: Russia is the world's leading natural gas exporter and second largest natural gas producer, \\ while also the second largest oil exporter and the third largest oil producer. \\
    \midrule
        Q: Who did gibbs shoot in season 10 finale / Gold: Mendez \\
        Predictions: fbi agent tobias fornell (from all systems in unrestricted/6Gb track) 
        \\
        {\em Past, Present and Future (NCIS)} Gibbs realizes that Mendez is .. The show then returns to the scene \\ depicted in a scene from season 10 finale's ... shoots and kills Mendez.  \\
    \midrule
        Q: The girl next door film based on a true story / Gold: loosely / Prediction: girl next door	\\
    \bottomrule
\end{tabular}
\caption{
    Samples of easy questions (all systems predict correct answers) and hard questions (all systems predict incorrect answers).
}
\label{tab:analysis-easy-hard}
\end{center}
\end{table}

\paragraph{Error Analysis}

We further present an error analysis of the top two systems in the \low{} track: \ucl{} and \naver{}. We choose these two systems because their approaches differ significantly from each other and the \low{} track fits the main motivation of the EfficientQA competition. We randomly sample 100 questions from the development data.
40, 33 and 27 questions are answered correctly by both systems, one of the systems,
and none of the systems, respectively.

Table~\ref{tab:analysis-ucl} shows the breakdown of the predictions from \textbf{\ucl}, where 50 questions are answered correctly, and the other 50 are not.
A majority of the error cases (47 out of 50) is due to retrieving an incorrect question.
Out of 47, 25 cases retrieve the question with a different topic from the input
questions, many of which discuss different entities
(e.g., ``don't it make you feel like dancing'' vs. ``you make me feel like dancing''). 
The other 19 cases retrieve the question that discusses the same topic or contains the same key entity, but is asking about different details. For instance, in the example in Table~\ref{tab:analysis-ucl}, the input question asks the amount {\em spent for} a film, while the retrieved question ask the amount {\em made by} the film.
It is also worth noting that the system sometimes gets the correct answer from the retrieved question with a different meaning from the input question, e.g., ``What republican is running for mayor of phoenix'' vs. ``Who is the current mayor of phoenix'', which have the same answer because the current mayor of Phoenix is Republican.

Table~\ref{tab:analysis-naver} shows the breakdown of the predictions from \textbf{\naver}, where 63 questions are answered correctly, and 37 are not. Failure due to pruning the gold passage is rare, being responsible for only 3\%. More of the failure cases are due to missing the gold passage, either when choosing 80 passages through dense retrieval (12\%), or during cross-attention re-ranking 80 passages to decide on the top 1 passage (15\%). Finally, in 7\% of the cases, the top 1 passage contains the gold answer but the system fails to extract the correct answer.
The gold passages in this category have valid but implicit support of the answer, e.g., the last example in Table~\ref{tab:analysis-naver}.

\begin{table}[t]
\begin{center} \setlength\tabcolsep{4pt} \scriptsize 
\begin{tabular}{ll}
    \toprule
        \multirow{6}{*}{Correct} & \textbf{Captured by automatic eval (32\%)} \\
        & Input: Who played apollo creed in the original rocky (A: carl weathers) \\
        & Retrieved: Who played apollo creed in the first rocky movie (A: carl weathers) \\
    \cline{2-2}
        & \textbf{Captured by manual eval (18\%)} \\
        & Input: When does the new season of snl air (A: September 29, 2018) \\
        & Retrieved: When does the new season of snl 2017 start (A: September 30, 2017) \\
    \midrule
        \multirow{15}{*}{Incorrect} & \textbf{Retrieved Q is not related to input Q in topics (25\%)} \\
        & Input: Who sings don't it make you feel like dancing \\
        & Retrieved: Who sings the song you make me feel like dancing \\
    \cline{2-2}
        & \textbf{Retrieved Q is related but is asking different detail (19\%)} \\
        & Input: How much did it cost to make bohemian rhapsody film \\
        & Retrieved: How much money did the movie rhapsody make \\
    \cline{2-2}
        & \textbf{Retrieved Q is related but contains incorrect `wh' word (3\%)} \\
        & Input: When was the current season of top chef filmed \\
        & Retrieved: What is the new season of top chef  \\
    \cline{2-2}
        & \textbf{Retrieved Q is context-dependent (1\%)} \\
        & Input: Who won the war of 1812 between russia and france \\
        & Retrieved: Who did we fight in the war of 1812 \\
    \cline{2-2}
        & \textbf{Retrieved Q has incorrect answer (2\%)} \\
        & Input: When was the last time florida state had a losing record \\
        & Retrieved: When was the last time florida state football had a losing season \\
    \bottomrule
\end{tabular}
\caption{
    Breakdown of the predictions from \ucl{} on 100 random samples.
}
\label{tab:analysis-ucl}
\end{center}
\end{table}

\begin{table}[t]
\begin{center} \setlength\tabcolsep{4pt} \scriptsize 
\begin{tabular}{ll}
    \toprule
        \multirow{7}{*}{Correct} & \textbf{Captured by automatic eval (34\%)} \\
        &Q: Who played apollo creed in the original rocky / Gold: Carl Weathers / Prediction: Carl Weathers \\
        & Retrieved P: {\em Carl Weathers}: He is best known for portraying apollo creed in the ``rocky'' series of films. \\
    \cline{2-2}
        & \textbf{Captured by manual eval (29\%)} \\
        & Q: what was the shelby in gone in 60 seconds / Gold: Shelby Mustang GT500 / Prediction: gt500 \\
        & Retrieved P: {\em Eleanor}: The eleanor name is reused for a shelby mustang gt500 in the 2000 gone in \\
        & 60 seconds remake. \\
    \toprule
        \multirow{19}{*}{Incorrect} & \textbf{No answer found in pruned Wiki (3\%)} \\
        & Q: Who has been chosen for the 2017 saraswati samman \\
        & Gold: Sitanshu Yashaschandra / Prediction Vijayendra saraswathi \\
        & Retrieved P: {\em jayendra saraswathi}: The mutt’ s pontiff sri vijayendra saraswathi performed the poojas \\
        & for his guru and predecessor. \\
        & Gold P: {\em Sitanshu Yashaschandra}: He received Saraswati Samman (2017) for his poetry collection.. \\
    \cline{2-2}
        & \textbf{Fail to retrieve the gold passage (12\%)} \\
        & Q: Who sings don't it make you feel like dancing / Gold: the Headpins / Prediction: leo sayer \\
        & Retrieved P: {\em you make me feel like dancing} is a song by the british singer leo sayer .. \\
    \cline{2-2}
        & \textbf{Fail to rerank the gold passage (15\%)} \\
        & Q: Who says all animals are equal but some are more equal than others \\
        & Gold: The pigs / Prediction: aristotle \\
        & Retrieved P: {\em moral status of animals in the ancient world}: Aristotle perceived some similarities \\
        & between humans and other species and developed a sort of ``psychological continuum'', recognising \\
        & that human and non-human animals differ only by degree in possessing certain temperaments.  \\
    \cline{2-2}
        & \textbf{Fail to extract the correct answer (7\%)} \\
        & Q: Which is the smallest continent in size in the world / Gold: Australia / Prediction: Asia \\
        & Retrieved P: {\em Continent} ...Ordered from largest in area to smallest, they are: Asia, Africa, North \\
        & America, South America, Antarctica, Europe, and Australia \\
    \bottomrule
\end{tabular}
\caption{
    Breakdown of the predictions from \naver{} on 100 random samples.
}
\label{tab:analysis-naver}
\end{center}
\end{table}

\white{.}

\ifsqueeze

\fi

\end{document}